\documentclass[12pt]{article}
\usepackage[a4paper, total={6in, 8in}]{geometry}

\pdfoutput=1
\usepackage[utf8]{inputenc}
\usepackage[USenglish]{babel}

\usepackage{fancyhdr}
\usepackage{microtype}
\usepackage{graphicx}
\usepackage{subfigure}
\usepackage{booktabs}
\usepackage{amssymb,amsthm,amsfonts}%amsmath
\usepackage{MnSymbol}
\usepackage{mathrsfs}
\usepackage{slashed}
\usepackage{color,rotating}
\usepackage{dsfont}
\usepackage{setspace}
\usepackage{physics}
\usepackage{verbatim}
\usepackage{hyperref,doi}
\usepackage[export]{adjustbox}
\usepackage[normalem]{ulem}
\usepackage{bbold}
\usepackage[linesnumbered,ruled,vlined]{algorithm2e}
\usepackage{algpseudocode}

\usepackage{authblk}

\SetKwInput{KwInput}{Input}    
\SetKwInput{KwOutput}{Output}

\usepackage{graphicx}

\usepackage[labelfont=bf, font=footnotesize]{caption}

\makeatletter
\def\l@subsubsection#1#2{}
\makeatother

\begin{document}

\title{Closed-Form Interpretation of Neural Network Classifiers with Symbolic Gradients}

\author{Sebastian J. Wetzel}
\affil{University of Waterloo, Waterloo, Ontario N2L 3G1, Canada}
\affil{Perimeter Institute for Theoretical Physics, Waterloo, Ontario N2L 2Y5, Canada}
\affil{Homes Plus Magazine Inc., Waterloo, Ontario N2V 2B1, Canada}

\affil{\texttt{swetzel@perimeterinstitute.ca}}
\date{}

%%%%%%%%%%%%%%%%%%%%%%%%%%%%%%%%
\maketitle
\begin{abstract}
I introduce a unified framework for finding a closed-form interpretation of any single neuron in an artificial neural network. Using this framework I demonstrate how to interpret neural network classifiers to reveal closed-form expressions of the concepts encoded in their decision boundaries. In contrast to neural network-based regression, for classification, it is in general impossible to express the neural network in the form of a symbolic equation even if the neural network itself bases its classification on a quantity that can be written as a closed-form equation. The interpretation framework is based on embedding trained neural networks into an equivalence class of functions that encode the same concept. I interpret these neural networks by finding an intersection between the equivalence class and human-readable equations defined by a symbolic search space.  The approach is not limited to classifiers or full neural networks and can be applied to arbitrary neurons in hidden layers or latent spaces.
\end{abstract}
\vspace{2pc}

%%%%%%%%%%%%%%%%%%%%%%%%%%%%%%%% 

\newpage

%\section{TL;DR}
% What do you do in practice? To interpret an artificial neural network, I fit the normalized gradients of a symbolic regression model to the normalized gradients of the output neuron of the neural network.

% I propose to interpret an artificial neural network by performing symbolic regression on the gradients of an output neuron with respect to the data. Through this procedure, one can find functions that base their decisions on the same quantity as the neural network but written in symbolic form.
% \\
% Why are you doing that? To interpret a neural network, one is, in general, not interested in finding a symbolic form of a neural network, but in finding a human-readable representation of the concept encoded in the neural network. Hence, it is helpful to get rid of uninterpretable transformations to gain access to a human-readable form of the underlying concept.

\section{Introduction}
Artificial neural networks have become the most powerful tool in a wide range of machine learning topics such as image classification, speech recognition, or natural language processing. In many fields, artificial neural networks have achieved better than human performance, and the number of these fields is consistently growing. The larger and more powerful the artificial neural networks become the more elusive is their decision-making to us humans. The black-box nature of these models makes it very challenging to comprehend their decision-making processes. 

Interpreting artificial neural networks is a central endeavor in many applications. As artificial neural networks are entering domains where a detailed understanding of the decision-making process is crucial, it becomes increasingly important to interpret their inner workings. These fields contain safety-critical, medical \cite{Jin2022,Amann2022} applications where human lives could be in danger or the legislative domains \cite{Hacker2020,Bibal2020} in which each decision needs to be supported by law. Self-driving cars are in danger of causing accidents because of a lack of understanding of the underlying machine learning algorithms \cite{Veran2023}. Further, in many scientific applications \cite{Lusch2018,cranmer2020discovering,Lemos2023,Wetzel2017a,Miles2021,Wetzel2020,Ha2021,Liu2021,Oviedo2022} the primary utility of a model is to help scientists understand an underlying phenomenon much more than making accurate predictions.

There has been a lot of progress in identifying which input features contribute the most to a certain decision \cite{Bach2015,sundararajan2017axiomatic,shrikumar2017learning}. Using these approaches it is possible to identify where danger lingers in an image of a traffic situation or find malignant cancer cells in MRI scans. However, from this understanding of low-level features, it is still a long road to develop a comprehensive understanding of the concepts an artificial neural network learns in order to make its predictions. The review \cite{Montavon2018} provides an overview of explainable artificial intelligence, covering methods for understanding, visualizing, and interpreting deep learning models.

There are many examples, especially in artificial scientific discovery, where a low-level interpretation is not enough. In many scientific fields, it is imperative to gain a deep understanding of the concept that is underlying neural network predictions. Some of the most central scientific tasks are a) understanding the dynamics of systems b) explaining new phases of matter and c) finding conserved quantities and symmetry invariants. In recent years, many scientists have developed methods to approach these problems. There are methods for the automated discovery of the dynamics of a system \cite{Brunton2016}, symbolic regression for conserved quantities \cite{Schmidt2009}, and support vector machines that map out phase diagrams \cite{Ponte2017,Greitemann2019}. These methods are inherently interpretable. Similarly, much more powerful neural network approaches have been developed to address these problems, which tend to be more successful in learning the underlying concepts but conceal these concepts from a human scientist. However, there have been publications that report successful strategies to discover the underlying concepts from neural networks. These include understanding the equations behind a dynamical system \cite{Lusch2018,cranmer2020discovering,Lemos2023}, extracting phase signatures from a neural network \cite{Wetzel2017a,Miles2021} or revealing conserved quantities and symmetry invariants with neural networks \cite{Wetzel2020,Ha2021,Liu2021}. It is important to note that many of the above examples are tackled by neural network classifiers while most symbolic interpretation literature focuses on regression problems. 

To address the problem of interpreting neural network classifiers, it is crucial to acknowledge that concepts encoded in an artificial neural network are stored in a highly convoluted and elusive manner. A neural network does not encode them in a human-readable form, but through complex interactions of thousands if not millions of neurons/perceptrons. In other words, even if the neural network learns a concept that could be put into a human-readable form, a neural network hides this concept through complex and highly nonlinear transformations.

This observation leads to the initial idea behind the interpretation framework proposed in this manuscript. Is it possible to extract a human-readable concept $g$ that is learned by a neural network $F$ by getting rid of the elusive transformation $\phi$ that conceals it?
\begin{align}
F(x_1,...,x_n)=\underbrace{\text{sigmoid}}_{\text{activation function}}( \underbrace{\phi}_{\text{uninterpretable transformation}} ( \underbrace{g(x_1,...,x_n)}_{\text{closed form decision function} } ))
\end{align}
The current paper provides an answer to this question by embedding the neural network $F$ into an equivalence class of functions that learn the same concept. A closed-form expression of this concept can be recovered by finding the intersection of this equivalence class with the space of human-readable functions defined through a symbolic search space.

% \subsection{Outline}

% This article is structured in the following manner: After exploring the most closely related prior research on neural network interpretability in Section~\ref{chapter:Prior Work}:Prior Work, I describe the formal and mathematical framework that is necessary for the interpretation algorithm in Section~\ref{chapter:Theoretical Framework}: Theoretical Framework. Next, I explain the computational implementation of the interpretation algorithm in Section~\ref{chapter:Interpretation Algorithm}:Interpretation Algorithm. Finally, I perform experiments demonstrating interpretation results in Section~\ref{chapter:Experiments}:Experiments. The findings and potential follow-up research are summarized in Section~\ref{chapter:Conclusion}: Conclusion.

%\section{Prior Work}
%\label{chapter:Prior Work}
To the best of my knowledge, there are no successful approaches that manage to find closed-form expressions of the high-level concepts learned by arbitrary neurons within a neural network. Much research has been done to find symbolic solutions to regression problems of which some can be extended to classification problems.

The most closely related works that also embed the neural network into another function space are about symbolic metamodels\cite{alaa2019demystifying,Abroshan2023,crabbe2020learning}, where the authors reproduce the output of an artificial neural network by minimizing the difference between the neural network output and so-called Meijer G-functions. However, this approach has so far only been successfully employed for low-dimensional regression problems.

It was proposed to train a neural network to map the weights of a neural network to a feature vector representing symbolic functions \cite{Bartelt2020}.

While their training efficiency and accuracy cannot compete with artificial neural networks, it is possible to directly employ symbolic regression approaches to obtain a symbolic solution that replicates the behaviour of a neural network. Many of these algorithms can be extended to classification problems through margin or hinge loss functions. Common symbolic regression libraries that mostly employ genetic algorithms include Eureqa \cite{Schmidt2009},  Operon C++\cite{Burlacu2020}, PySINDy \cite{Kaptanoglu2022}, Feyn\cite{Brolos2021}, Gene-pool Optimal Mixing Evolutionary Algorithm \cite{Virgolin2021}, GPLearn \cite{Stephens2022} and PySR\cite{Cranmer2023}. 

There are also symbolic regression algorithms that employ neural networks, for example, EQL with embedded symbolic layers\cite{Martius2016,sahoo2018learning,Dugan2020}, deep symbolic regression uses recurrent neural networks \cite{petersen2020deep}, and symbolic regression with transformers\cite{kamienny2022end,pmlr-v139-biggio21a}. Further, AI Feynman uses neural networks to simplify expressions \cite{Udrescu2020}. 

An overview of interpretable scientific discovery with symbolic Regression can be found in\cite{Makke2022}.

It is important to note that the process of replicating the output of a neural network faces a fundamental flaw: there is no guarantee that this procedure captures the same features that a neural network learns to solve a specific problem. Hence, replicating the output of a neural network is different from interpreting a neural network. This observation is demonstrated in section \ref{chapter:Experiments2}.

\section{Framework Overview}
\label{chapter:Theoretical Framework}

\begin{figure*}[h!]
    \centering
    \includegraphics[width=\textwidth]{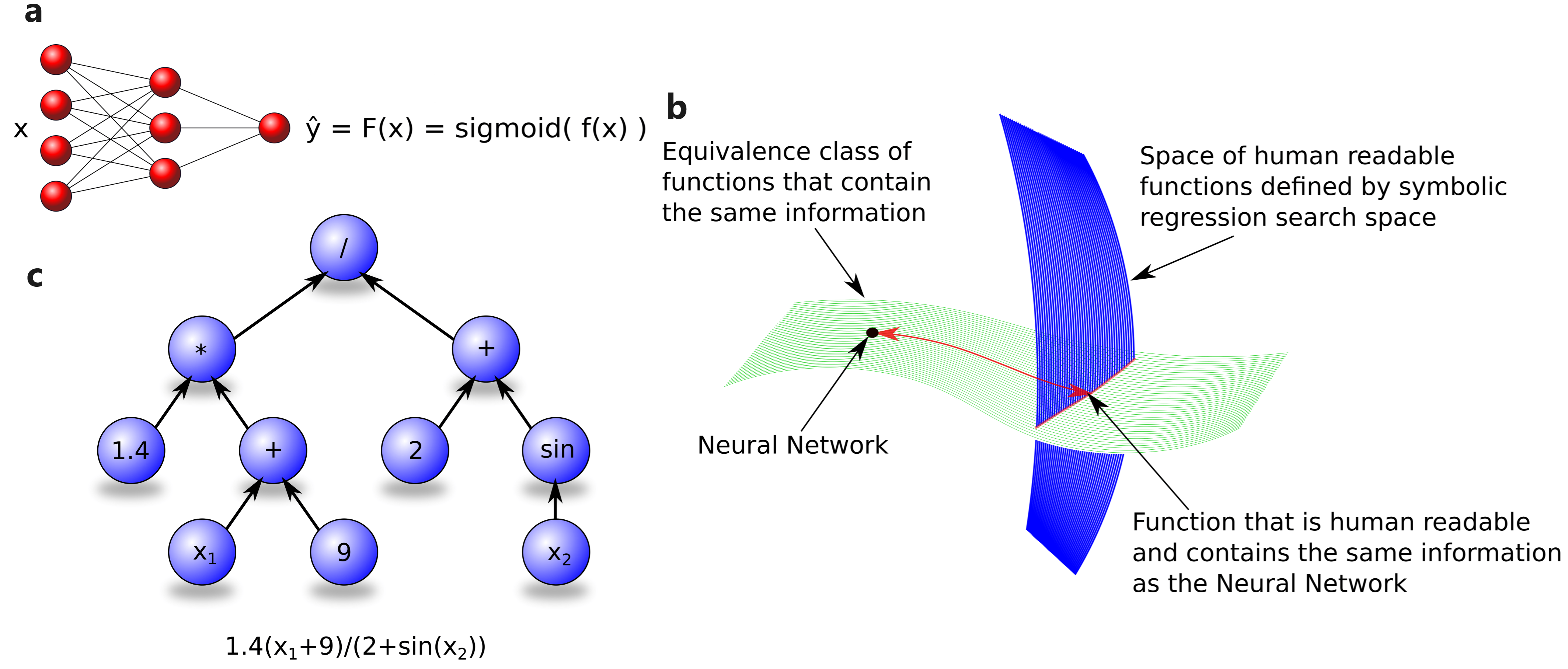}
    \caption{\textbf{a:} An artificial neural network is a connected graph consisting of nodes representing neurons and weighted connections between them. A neural network predicts an approximate target $F(\mathbf{x})=\hat y \approx y$ here in the context of binary classification. Removing the final sigmoid activation function allows the extraction of the latent model $f$ from the full neural network $F$ for easier interpretability.
    \textbf{b:} The interpretation framework is based on finding the intersection between human-readable functions and the equivalence class \ref{eq:H_g} of functions that contain the same information as an output neuron of the neural network. The space of human-readable functions can be defined through a symbolic search space with elementary functions and complexity that matches the user's knowledge. This space can be computationally explored by genetic algorithms whose structure is mathematically represented by \textbf{c:} a symbolic tree. A tree consists of connected nodes containing variables, numeric parameters, unary and binary operators. Symbolic search is performed by a genetic algorithm that modifies, evolves and adds nodes to optimize an objective function on some underlying training data.    
    }
    \label{fig:overview}
\end{figure*}

\subsection{Interpreting an Artificial Neural Network}
Interpreting an artificial neural network means formulating a mapping between an abstract mechanism or encoding into a domain that a human can understand. Human understanding is a vaguely defined concept since human capabilities range from simply perceiving common objects in the real world to expert knowledge in their respective field. I suggest categorizing interpretations along the following four different attributes, which is more fine-grained than the categories proposed in \cite{Montavon2018}:

A) Mechanistic vs. Functional: Mechanistic interpretation is concerned with explaining the mechanisms the neural network employs to solve a problem or to implement a function. Functional interpretation is about understanding the concepts by which a neural network relates the input to the output.

B) Local vs. Global: Local interpretation is about understanding which elements of a certain data point influence the prediction, while global interpretation explains what features are relevant for a learned concept in general.

C) Verify vs. Discover: A machine learning practitioner might have a set of hypotheses of concepts in mind that a neural network might learn in order to solve a given task. It is often easy to verify or falsify if these concepts are included in the features learned by an artificial neural network. However, discovering an unknown concept without a list of potential hypotheses is in general very hard.

D) Low-Level vs. High-Level Features: The neurons of the network jointly implement a complex nonlinear mapping from the input to the output through potentially several hidden layers. The global concept that must be interpreted is usually represented by a neuron in the final layer. Low-level features that are close to the input are usually rather interpretable while concepts in the final layers or some latent space are in general very abstract.

Interpreting an artificial neural network through the framework presented in this paper falls into the category of functional global discovering high-level interpretation.

\subsection{Equivalence Class of Functions Containing the Same Information}
The goal of the interpretation framework outlined in this manuscript is to find the intersection of the equivalence class of functions that contain the same information as the output neuron (or any other neuron) and the set of human-readable equations, see fig.~\ref{fig:overview}\textbf{b}.

For this purpose, it is important to define what it means that a certain neuron in an artificial neural network learns to encode a certain quantity. This neuron can be seen as a function $f(\mathbf{x})$ that depends on the input $\mathbf{x}$. $f(\mathbf{x})$ contains the full information about a certain quantity $g(\mathbf{x})$ if $g(\mathbf{x})$ can be faithfully reconstructed from $f(\mathbf{x})$. Conversely, if $f(\mathbf{x})$ only contains information from $g(\mathbf{x})$ it is possible to reconstruct $f(\mathbf{x})$ from the knowledge of $g(\mathbf{x})$. In mathematical terms that means that there exists an invertible function $\phi$ such that $f(\mathbf{x}) = \phi(g(\mathbf{x}))$. 

For the purpose of this work, $g(\mathbf{x}),f(\mathbf{x}) \in C^1(D \subset \mathbb{R}^n,\mathbb{R})$, are assumed to be continuously differentiable functions, where $C^1(D\subset\mathbb{R}^n,\mathbb{R})$ denotes the vector space of differentiable functions from $D$ to $\mathbb{R}$. Further, $D\subset\mathbb{R}^n$ is the data manifold which is required to be compact and simply connected. By employing these definitions it is possible to define an equivalence set of all functions that extract the same scalar-valued information from the data:   
\begin{align}
\tilde H_g=\left\{ f(\mathbf{x}) \in C^1(D \subset \mathbb{R}^n,\mathbb{R}) \ | \exists \text{ invertible } \phi \in C^1(\mathbb{R},\mathbb{R}) \ : \ f(\mathbf{x}) = \phi(g(\mathbf{x}))\right\} \label{eq:H_gtilde}
\end{align}
If $f(\mathbf{x})$ is the output of an artificial neural network, this defines an equivalence relation between all functions that base their decision on the same quantity $g(\mathbf{x})$. Several different neural networks might encode the same function $f$ because, in the context of this section, it does not matter how the output is calculated mechanistically. All realizations of a neural network that is symmetric under permutations or other changes in its weights would still be functionally equivalent and thus the same function within the equivalence class $\tilde H_g$.

Since it is computationally very inefficient to verify whether two functions or neural networks belong to the equivalence class $\tilde H_g$, according to the above definition, it is convenient to reformulate it. Taking the gradient of a function $f\in \tilde H_g$ with respect to the input yields
\begin{align}
\nabla f(\mathbf{x}) = \phi'(g(\mathbf{x})) \nabla g(\mathbf{x}) 
\end{align}
Since $\phi$ is invertible, $\phi'$ is strictly monotonic, thus $|\phi'(g(\mathbf{x}))|>0$, hence the gradients of $f(\mathbf{x})$ and $g(\mathbf{x})$ are always parallel $\nabla f(\mathbf{x})|| \nabla g(\mathbf{x})$ at every $\mathbf{x}\in D$.

Using this property, I define the equivalence set $H_g$, with $ \tilde H_g\subseteq H_g$, which will later form the basis for the neural network interpretation algorithm:
\begin{align}
H_g=\left\{ f(\mathbf{x}) \in C^1(D \subset \mathbb{R}^n,\mathbb{R}) \ | \ \frac{\nabla f(\mathbf{x})}{\lVert \nabla f(\mathbf{x})\rVert} = \frac{\nabla g(\mathbf{x}) }{\lVert \nabla g(\mathbf{x})\rVert} \lor \nabla f(\mathbf{x})=\nabla g(\mathbf{x})=0, \ \forall \mathbf{x} \in D \right\} \label{eq:H_g}
\end{align}
Here $ \lVert \cdot \rVert $ is the Euclidean norm used to normalize the gradients to unit length. The equivalence classes $\tilde H_g$ and $H_g$ satisfy the definition of equivalence classes. $\forall f_1,f_2,f_3 \in \tilde H_g \text{ or } H_g$: a) Reflexivity: $f_1 \sim f_1$ , b) Symmetry: $f_1 \sim f_2$ implies $f_2 \sim f_1$ , Transitivity: if $f_1 \sim f_2$ and $f_2 \sim f_3$ then $f_1 \sim f_3$. Trivially, if $f\in H_g$ then $H_f=H_g$. It can be proven by using the above definitions that $H_g=\tilde H_g$.

%  If $\forall x \in D: \nabla f(x)\neq 0$ and $\nabla g(x)\neq 0$ and thus $D=D_0$ then 
% as long as $\nabla f(x)\neq 0$ and $\nabla g(x)\neq 0$. By defining $D_0\subset D \setminus \{ x | \nabla f(x)\neq 0 \lor \nabla f(x)\neq 0 \}$

\subsection{Equivalence of Equivalence Classes}
\subsubsection{Proof}
\label{chapter:proof}
Let $g(\mathbf{x}),f(\mathbf{x}) \in C^1(D \subset \mathbb{R}^n,\mathbb{R})$ be continuously differentiable functions ($C^1(D\subset\mathbb{R}^n,\mathbb{R})$ is the vector space of differentiable functions from $D$ to $\mathbb{R}$) and $D\subset\mathbb{R}^n$ be the data manifold which is required to be compact and simply connected. Then $\tilde H_g= H_g$ where 
\begin{align}
\tilde H_g=\left\{ f(\mathbf{x}) \in C^1(D \subset \mathbb{R}^n,\mathbb{R}) \ | \exists \text{ invertible } \phi \in C^1(\mathbb{R},\mathbb{R}) \ : \ f(\mathbf{x}) = \phi(g(\mathbf{x}))\right\} 
\end{align}
and
\begin{align}
H_g=\left\{ f(\mathbf{x}) \in C^1(D \subset \mathbb{R}^n,\mathbb{R}) \ | \ \frac{\nabla f(\mathbf{x})}{\lVert \nabla f(\mathbf{x})\rVert} = \frac{\nabla g(\mathbf{x}) }{\lVert \nabla g(\mathbf{x})\rVert} \lor \nabla f(\mathbf{x})=\nabla g(\mathbf{x})=0, \ \forall \mathbf{x} \in D \right\} 
\end{align}

Proof: One can see that for each function $f\in \tilde H_g \ \phi:\nabla f(\mathbf{x}) = \phi'(g(\mathbf{x})) \nabla g(\mathbf{x})$ , hence $\tilde H_g \subset H_g$.

It remains to be shown that for each function $f\in H_g \ \exists \phi : f(\mathbf{x})=\phi(g(\mathbf{x}))$. It is possible to explicitly construct the function $\phi$ that maps between $f$ and $g$. Defining $\phi'$ through 
\begin{align}
\nabla f(\mathbf{x}) = \phi'(g(\mathbf{x})) \nabla g(\mathbf{x})\label{eq:grad}
\end{align}
leads to an integrable $\phi'(g(\mathbf{x})) \nabla g(\mathbf{x}) = \nabla f(\mathbf{x})$ because a) the images of $f(D)$ and $g(D)$ are compact, thus $\phi'$ maps between compact subsets of $\mathbb{R}$ and b) $\phi'$ is continuous. For any simply connected $D \subset \mathbb{R}^n$ I can define the $C^1$-curve $\mathbf{x}:[t_0,t_1]\rightarrow D$, thus a variable transformation within the calculation of the contour integral yields:

\begin{align}
\phi(g(\mathbf{x}(t_1)))-\phi(g(\mathbf{x}(t_0)))&=\int_{g(\mathbf{x}(t_0))}^{g(\mathbf{x}(t_1))} \phi'(g)\ dg \\
&=\int_{\mathbf{x}(t_0)}^{\mathbf{x}(t_1)} \phi'(g(\mathbf{x})) \nabla g(\mathbf{x})\cdot d\mathbf{x} \\
&=\int_{t_0}^{t_1} \phi'(g(\mathbf{x}(t))) \nabla g(\mathbf{x}(t)) \cdot \dot{ \mathbf{x}}(t) \ dt \\
&\stackrel{eq.\ref{eq:grad}}{=}\int_{t_0}^{t_1}  \nabla f(\mathbf{x}(t))  \cdot \dot{ \mathbf{x}}(t) \ dt \\
&=f(\mathbf{x}(t_1)) - f(\mathbf{x}(t_0))
\end{align}
Note, since I base the integral on the product $\phi'(g(\mathbf{x})) \nabla g(\mathbf{x})$ instead of $\phi'(g(\mathbf{x}))=\lVert \nabla f(\mathbf{x})\rVert /\lVert \nabla g(\mathbf{x})\rVert$, I avoid problems arising from $\nabla f(\mathbf{x})=\nabla g(\mathbf{x})=0$. Similarly, one can proof the existence of $\tilde \phi:\tilde \phi(f(\mathbf{x}))=g(\mathbf{x})$ such that $f(\mathbf{x})=\phi(\tilde\phi(f(\mathbf{x})))$ and thus $\phi$ is invertible. Having explicitly constructed $\phi$ proofs $H_g=\tilde H_g$.

\subsubsection{Assumptions}
In a practical machine learning applications not all assumptions from the prior section that ensure $H_g = \tilde H_g$ hold true. However, even then $H_g \approx \tilde H_g$ can provide a good approximation that allows for a retrieval of the function that a neuron encodes.

A machine learning data set $D \subset \mathbb{R}^n$ is bounded since it is finite. However, if there is a divergence in the function that the machine learning model is supposed to approximate, the data set might not be closed and thus not compact. A data set $D$ might not be simply connected, especially if it is in the form of categorical data or images.

A neural network classifier, if successfully trained, tends to approximate a categorical output, which is neither continuous nor differentiable. However, this binary output is typically an approximation mediated by sigmoid or softmax activation functions, which indeed are continuously differentiable. Still, interpreting artificial neural networks with the framework introduced in this paper experiences numerical artifacts if a gradient is taken from a network that contains sigmoid or softmax activation functions. For this reason, I suggest avoiding these activation functions in the design of hidden layers and removing them from the output neuron during the interpretation process (the same argument holds true for tanh or related activation functions).

The above definitions of equivalence classes could be extended to piecewise $C^1(\mathbb{R}^n,\mathbb{R})$ functions. This function set contains many artificial neural networks that include piecewise differentiable activation functions like $\text{ReLU}(x)=\max(0,x)$. However, this causes problems when evaluating derivatives close to $\text{ReLU }(x)=0$. In practice, one can observe that piecewise $C^1$ activation functions lead to computational artifacts when calculating gradients. Hence, I suggest using $\text{ELU}=\exp(x)-1 |x\leq 0, x | x > 0$ as the preferred activation function in hidden layers.

\subsection{Extracting Symbolic Concepts Encoded in Neural Networks}
A neural network classifier $F$ can be interpreted in practice by finding a representative symbolic function that lives on the intersection between human-readable functions and all functions that contain the same information as a trained neural network, see fig.~\ref{fig:overview}\textbf{b}. For this purpose, I employ a genetic algorithm to find a function represented by a symbolic tree $T$ whose normalized gradients approximate the normalized gradients of a latent neural network model $f$. Here $f$ is defined by removing the final activation function $F(x)=\text{sigmoid} (f(x))$. Through this procedure it is possible to find an element belonging to the equivalence class \ref{eq:H_g} in human-readable form.

While the experiments here focus on interpreting binary classifiers, it is possible to apply the framework to interpret multi-class classifiers or even neurons contained in the hidden layers of neural networks. The interpretation framework contains three algorithmic steps summarized in algorithm \ref{algcollection}: 1) training an artificial neural network, 2) extracting gradients of the output neuron with respect to the input and 3) performing symbolic search to find a symbolic function whose normalized gradients approximate the normalized gradients from the desired neuron of an artificial neural network.

\section{Methods}

\subsection{Artificial Neural Network}

An artificial neural network fig.~\ref{fig:overview}\textbf{a} is a connected graph containing nodes representing neurons otherwise known as perceptrons and connections representing weighted inputs that are supplied to these nodes. A basic feedforward neural network can be represented mathematically as follows:

A neural network $F$ is applied to an input data point $ \mathbf{a}^{(0)} = \mathbf{x} \in \mathbb{R}^n$ represented through an embedding in an $n$-dimensional space of real numbers. The input gets processed through a number $l$ of layers $L$ containing weight matrices $\mathbf{W} \in \mathbb{R}^{m_l\times m_{l-1}}$, biases vectors $\mathbf{b} \in \mathbb{R}^{m_l}$ and non-linear activation functions $\sigma$
\begin{align}
\mathbf{a}^{(l)}=L^{(l)}(\mathbf{a}^{(l-1)})= \sigma^{(l)}(\mathbf{W}^{(l)} \mathbf{a}^{(l-1)} + \mathbf{b}^{(l)})
\end{align}
to produce some output
\begin{align}
\mathbf{\hat y} = F(\mathbf{x})=(L^{(l)}\circ ... \circ L^{(0)})(\mathbf{x})
\end{align}
The parameters of an artificial neural network are trained on a data set $(X,Y)$ representing the data manifold $X\subset D \subset \mathbb{R}^n$ through gradient descent/backpropagation to minimize an objective/loss/error function $\mathcal{L}(F(X),Y)$ such that $F(X)\approx Y$ on all data points on the data manifold.

\subsection{Symbolic Search}
\subsubsection{Symbolic Search Space}

An interpretable representation of a neural network is a description that humans can make sense of. Each single neuron can be described as a simple linear mapping together with an activation function that can be expressed in terms of a short equation. However, the complexity that arises from combining several thousand or even millions of these neurons conceals any meaning of the function that is approximated by a neural network. In order to discuss the requirements for interpretable representations in the language of mathematics, let me discuss three criteria that are important for comparing representations of equations.

1)Building blocks: In the realm of mathematical equations human readable equations are typically written in terms of numbers (integers, real numbers, complex numbers), binary ($+$,$-$,$\times$) operations, or unary operations that summarize complex operations or arise from the solutions of ubiquitous equations ($d/dx$,$\int \cdot\ dx$,$\exp(x)$,$\sin(x)$,$x!$). 

2) Complexity: Humans can understand short combinations of the above elementary constituents of equations, however, combining a large number of them prevents interpretability. Hence, to formulate complex objects, it is helpful to summarize and simplify repeating processes in equations to a shorter form. For example, the Lagrangian of the standard model of particle physics \cite{Woithe2017} fits on a coffee mug in its most simple formulation. However, each element in the equation represents a collection of more complex objects that are conveniently grouped together. 

3) Context: Pricing financial derivatives with the Black-Scholes equation \cite{Black1976} requires a firm understanding of the assumptions made to derive it. Identifying something as a Lagrangian or a conserved quantity means associating a certain functional purpose or properties to an equation. For example, reading the standard model Lagrangian in its full glory is a futile exercise, unless the reader has a deep understanding of what role Lagrangians play in the larger framework of quantum field theory. The same can be said about much simpler equations like the de-Broglie wavelength $\lambda=h/p$ ($h$-Panck constant, $p$-momentum) \cite{Broglie1924}, which is arguably one of the simplest physics equations. This equation is meaningful because of the context of its formulation, it is instrumental in associating a quantum mechanical wavelength to massive objects.

Throughout this paper, I base the set of human-readable functions on symbolic trees, see fig.~\ref{fig:overview}\textbf{c}. These trees can be constructed to contain a user-defined set of elementary operations. Symbolic trees can also be associated with a complexity measure that increases the larger the tree grows leading to several solutions along the Pareto front. Hence, it is possible to address human readability according to aspects 1) and 2). Within the scope of this paper, it is impossible to address the question of the context that gives the equation meaning, as discussed in 3). In recent years, progress in artificial scientific discovery has made it possible to train neural networks to optimize loss functions based on concepts that imply a context, instead of optimizing for regression targets \cite{Wetzel2017a,Wetzel2020,Liu2021,Ha2021}. The interpretation framework developed in the current manuscript is uniquely suitable to address the interpretation in these cases.

\subsubsection{Symbolic Search Algorithm}

In order to computationally search through the space of human-readable functions and determine the intersection with the equivalence set described in the previous section, it is convenient to build upon a suitable symbolic regression algorithm based on a backend that performs an efficient search in the space of functions represented by symbolic trees.

In principle, a simple way of searching through the space of possible functions could be optimizing a linear combination of user-defined elementary functions. A much better approach can be taken by utilizing evolutionary or genetic algorithms. These algorithms build trees of connected nodes that represent a mathematical function, see fig.~\ref{fig:overview}\textbf{c}. Each node represents input variables, numeric parameters, as well as unary and binary operators. A genetic algorithm modifies, evolves, and adds nodes to optimize an objective function on some underlying training data.

Many such algorithms have been developed in recent years, these include: Eureqa \cite{Schmidt2009},  Operon C++ \cite{Burlacu2020},  PySINDy \cite{Kaptanoglu2022}, Feyn \cite{Brolos2021}, Gene-pool Optimal Mixing Evolutionary Algorithm \cite{Virgolin2021} and GPLearn \cite{Stephens2022}. Other symbolic regression techniques based on neural networks are EQL \cite{Martius2016,sahoo2018learning}, AI Feynman \cite{Udrescu2020}, Deep symbolic regression \cite{petersen2020deep} and Symbolic regression with transformers \cite{kamienny2022end}.

In this paper, I build upon PySR \cite{Cranmer2023}, a high-performance symbolic regression algorithm for Python. PySR’s internal search algorithm is a multi-population genetic algorithm, which consists of an evolve-simplify-optimize loop, designed for the optimization of unknown scalar constants in newly-discovered empirical expressions. In order to optimize the computation speed, PySR’s backend is coded in Julia as the library SymbolicRegression.jl.

\subsection{Interpretation Algorithm}
\label{chapter:Interpretation Algorithm}

\begin{figure*}[h!]
    \centering
    \includegraphics[width=1\textwidth]{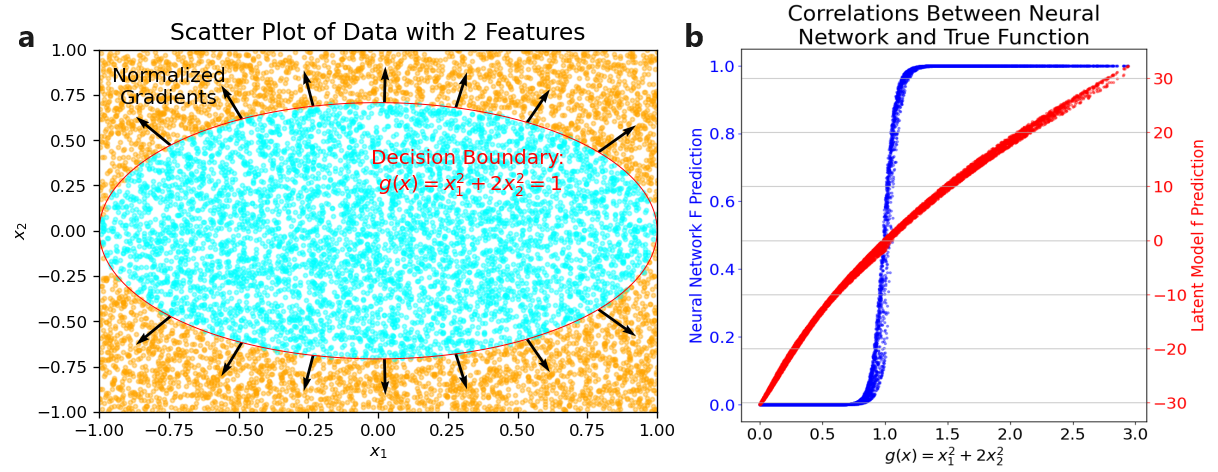}
    \caption{\textbf{a:}  Two class data of Experiment 1 separated by decision boundary $g(\mathbf{x})=x_1^2+2x_2^2=1$. A neural network $F$ is trained to classify the data. Afterward, a symbolic model $T$ is trained to reproduce the normalized gradients of $F$ which coincide with the normalized gradients of function that defines the decision boundary $g$. \textbf{b:} Empirical correlation between true function $g$ and the neural network $F$. Removing the sigmoid activation function from $F$ defines the latent model $f$ which has an almost linear correlation with $g$. However, this correlation is not linear and defines the function $f=\phi(g)$ with which I ascertain the equivalence relation $f\sim g$ assuring that $F$ and $f$ contain the same information as $g$ and thus $F,f\in\tilde H_g$ according to eq.\ref{eq:H_gtilde}.}
    \label{fig:data_prediction}
\end{figure*} 

\subsubsection{Training a Neural Network For Binary Classification}
The first part of the interpretation procedure involves training an artificial neural network for binary classification on a two-class data set, see algorithm \ref{algcollection}.1, to approximate the decision boundary that separates the two classes, see fig.~\ref{fig:data_prediction}\textbf{a}. After successful training I obtain the full model $F$ and the latent model $f$ by removing the final sigmoid activation function, which has a cleaner correlation to a closed-form expression of the decision boundary $g$, see fig.~\ref{fig:data_prediction}\textbf{b}:
\begin{align}
F(\mathbf{x})=\text{sigmoid}(f(\mathbf{x}))
\end{align}
I implemented one neural network architecture for all experiments using the Python library tensorflow \cite{tensorflow2015-whitepaper}. This neural network consists of two hidden layers with 1000 neurons each and ELU activation functions. I compared different activation functions and ELU activations yielded gradients least affected by computational artifacts. The final layer contains one single neuron together with a sigmoid activation function, commonly used for binary classification. The weights and biases in the hidden layers are regularized with an $L^2$ penalty of $10^{-4}$. Further, I employ dropout regularization after each hidden layer with a chance of $20 \%$. I train the neural networks using the Adam optimizer and use learning rate decay callbacks that reduce the learning rate by 50\% loss stops decreasing. Further early stop callbacks stop the training process when converged. For this reason, it is enough to set the number of epochs large enough such that the early stopping is always triggered. The batch size in all cases is 100. I train on $80\%$ of the data set and use another 20\% for validation. Since I am not interested in calculating any test accuracy here, the validation and test set are the same.

\begin{algorithm}
\caption{\textbf{Neural Network Interpretation Algorithm}}\label{algcollection}
\begin{algorithmic}
\State
\end{algorithmic}
\textbf{1) Train Neural Network for Binary Classification}

\begin{algorithmic}[1]
    \Statex  \textbf{Data:} Labelled data set $D=(X_{train},Y_{train})$
    \Statex  \textbf{Input:} Neural Network Hyperparameters   
    \State Initialize neural network classifier with sigmoid activation at output $F$
    \State Train $F$ on $D$
    \State $f$ $\leftarrow$ remove sigmoid activation of output neuron of $F$   
    \Statex  \textbf{Output:} Trained model $F$, latent model $f$  
\end{algorithmic}
\vspace{0.2cm}

\textbf{2) Obtain Gradients of Latent Model}
\begin{algorithmic}[1]
    \Statex \textbf{Data:} Labelled data set $D=(X_{train},Y_{train})$
    \Statex \phantom{\textbf{Data:}} Unlabelled data set $D_u=(X_{u})$ \Comment{Artificial data not used for training}
    \Statex \textbf{Input:} Trained model $F$ and latent model $f$
    \Statex \phantom{\textbf{Input:}} Selection threshold $\delta$
    
    \State $\tilde{X}\leftarrow(X_{train},X_u)$  \Comment{Add artificial unlabelled data}
    \State  $\tilde{X}\leftarrow\tilde{X}\textbf{ where }F(\tilde{X})\in[\delta,1-\delta] $  \Comment{Select data close to decision boundary}
    \State $G_f\leftarrow[ \nabla f(\mathbf{x}) \text{ for } \mathbf{x} \text{ in } \tilde{X}]$ \Comment{Gradients of latent model wrt. input}
    \State $G_f\leftarrow  [\textbf{if }\nabla f(\mathbf{x})\neq0: \nabla f(\mathbf{x})/ \lVert \nabla f(\mathbf{x}) \rVert$ \Statex \phantom{aaaaa}$\textbf{ else }\nabla f(\mathbf{x}) \text{ for } \nabla f(\mathbf{x})  \text{ in } G_f$] \Comment{Normalize Gradients}
    
    \Statex  \textbf{Output:} $(\tilde X ,G_f)$
\end{algorithmic}
\vspace{0.2cm}

\textbf{3) Symbolic Search}
\begin{algorithmic}[1]  

  \Statex \textbf{Data:} Gradient data set $(\tilde X ,G_f)$
  \Statex  \textbf{Input:} Symbolic Search Hyperparameters
  \Statex\phantom{\textbf{Input:}} Set of unary and binary operations.
\State  initialize symbolic model $T$
 \State evolve $T$ \textbf{with} ( 
  \State  \phantom{aaa} $G_T\leftarrow[ \nabla T(\mathbf{x}) \text{ for } \mathbf{x} \text{ in } \tilde{X}]$ \Comment{Gradients of symbolic model}
  \State   \phantom{aaa} $G_T\leftarrow  [\textbf{if }\nabla T(\mathbf{x})\neq0: \nabla T(\mathbf{x})/ \lVert \nabla T(\mathbf{x}) \rVert $\Statex \phantom{aaaaaaaaa}$\textbf{ else }\nabla T(\mathbf{x}) \text{ for } \nabla T(\mathbf{x}) \text{ in } G_T$] )\Comment{Normalize Gradients}
\State \phantom{aaa} to minimize MSE$(G_f,G_T)$

   \Statex  \textbf{Output:} Symbolic model $T$
   \end{algorithmic}

\end{algorithm}

\subsubsection{Obtaining Gradients of Latent Model}
The second step is described in algorithm \ref{algcollection}.2, which is used to collect the gradient information, see fig.~\ref{fig:data_prediction}\textbf{a}, for the symbolic search step. It involves potentially adding additional unlabelled data points to the training data set $X$, either from available unlabelled data, sampling from the data manifold, or by perturbing training data. This data set is denoted $\tilde X$. I omitted increasing the training data set in my experiments, but in data-scarce domains, this could lead to a significant improvement in the symbolic search results.

The sigmoid activation function prevents the training on data points $\mathbf{x}$ for which the neural network makes almost certain predictions of $F(\mathbf{x})\approx0$ or $F(\mathbf{x})\approx1$. In these cases, the sigmoid activation function is almost flat suppressing any gradient information. For this reason, I delete select data points for which $F(\mathbf{x})\nin[\delta,1-\delta]$ from $\tilde X$. In my experiments, I choose $\delta=0.0001$.

Afterward, I calculate the normalized gradients $G_f=\nabla f/\lVert \nabla f \rVert$ of the latent model $f$, with respect to the input variables, on the remaining data points. Finally, together with the inputs they get stored in the form of a labeled data set $(\tilde X, G_f)$ that is used as training data of the symbolic search algorithm.

\subsubsection{Symbolic Search}

The third step of the interpretation algorithm builds upon symbolic regression, see fig.~\ref{fig:overview}\textbf{c}, to find human-readable equations describing the decision boundary $g$. The algorithm is summarized in algorithm \ref{algcollection}.3. This step employs the backend of the Python symbolic regression module PySR \cite{Cranmer2023} coded in Julia called SymbolicRegression.jl.

The interpretation framework proposed in this paper has requirements beyond what symbolic regression algorithms typically do. Symbolic regression models are represented by trees $T$ that are trained on a labeled data set $(X,Y)$ to reproduce an output $Y$. However, the task in this manuscript is to perform symbolic regression on the normalized gradient $G_f=\nabla f/\lVert \nabla f \rVert$ of a specific neuron within the neural network. Thus, the gradient set $G_f$ takes on the role of the data set label $Y$. Further, the symbolic tree must be differentiable in order to calculate the normalized gradients of the tree  $G_T=\nabla T/\lVert \nabla T \rVert$ itself. Finally, the desired solution of the symbolic search procedure is not finding a closed form expression of the normalized gradient $\nabla f/\lVert \nabla f \rVert$, but the function $f$ itself. An additional limitation of SymbolicRegression.jl is its limitation to one-dimensional regression targets in the standard form.

The extremely customizable library SymbolicRegression.jl can be programmed to 1) accept gradient information as data set, 2) calculate gradients of a symbolic regression tree 3) normalize these gradients and 4) optimize a custom loss function that compares normalized gradients.

I choose the mean square error (MSE) loss function between latent model normalized gradient $G_f$ and normalized gradient of the symbolic tree $G_t$ as the most straight-forward choice of the objective function to ascertain that both $f$ and $T$ fall into the same equivalence class defined by eq.\ref{eq:H_g}. This loss function is equivalent to the cosine loss (CSL) based on the scalar product of two vectors.

\begin{align}
\text{MSE}(G_f,G_T)&= \frac{1}{n}\sum_{i=1}^n \left( \lVert G_{f,i}-G_{T,i} \rVert^2 \right)
\\
&=\frac{1}{n}\sum_{i=1}^n \left( \underbrace{\lVert G_{f,i}\rVert^2}_{1}+\underbrace{\lVert G_{T,i} \rVert^2}_{1} -2G_{f,i}\cdot G_{T,i}\right)
\\
&=2-2\text{CSL}(G_f,G_T)
\end{align}
The symbolic regression model is trained to find a closed-form function corresponding to the decision boundary using the following building blocks:
\begin{enumerate}
\item input variables $x_1,...,x_n$
\item floating-point constants
\item binary operators: +, -, *, /
\item unary operators: sin, exp
\end{enumerate}
The hyperparameters are a batch size of 25, a maximum number of epochs of 200, an early stop callback, and a maximum tree size of 30 nodes.

The output of the symbolic search algorithm is not just a single closed-form function, but a set of symbolic functions that balance complexity and accuracy along the Pareto front. More precisely, the result contains functions with low complexity and low accuracy, and more complex functions are added if the complexity can be justified by an increase in accuracy.

\section{Results}
\label{chapter:Experiments}

\begin{table}[h!]
  \caption{Data sets are created by randomly sampling data points from multidimensional uniform distributions that are divided into two classes to represent binary classification problems.
  %\td{update with latest results}
  }
  \centering
  \begin{tabular}{lcl}

   & Variables & Decision Formula \\
\midrule
    Experiment 1 & 2 &$x_1^2+2x_2^2>1$\\
    Experiment 2 & 2 &$x_1^2+3x_1x_2+2x_2^2>5$\\
    Experiment 3 & 3 &$x_1^2+\sin(x_2+x_3)>1$\\
    Experiment 4 & 3 &$\exp(x_1-x_2)-\pi x_3>1$\\
    Experiment 5 & 3 &$x_1\exp(-\frac{1}{2}(x_2^2+x_3^2))>\frac{1}{5}$\\
    Experiment 6 & 6 &$\frac{x_1x_2}{(x_3-x_4)^2+(x_5-x_6)^2}>1$\\ 
    \hline 
    Experiment 7 & 4 & $x_1x_2>1 \land x_3x_4>1$ vs $x_1x_2<1 \land x_3x_4<1$
  \end{tabular}
\label{tab:data}
\end{table}
\subsection{Data}
I apply the outlined interpretation procedure to several neural networks that are tasked with classifying data sets that are each grouped into two classes. The classes are separated through a domain boundary defined by a decision formula $g$ and each data point is obtained by uniformly sampling within ranges that contain the decision boundary. The interpretation framework attempts to recover these functions from the neural networks.
I perform experiments on different data sets outlined in table \ref{tab:data}. For each data set, I sample 10000 data points and categorize them into classes according to the decision formulas. In all cases, the data sets are created with multiplicative Gaussian noise of $1\%$.

\begin{figure*}[h!]
    \centering
    \includegraphics[width=1\textwidth]{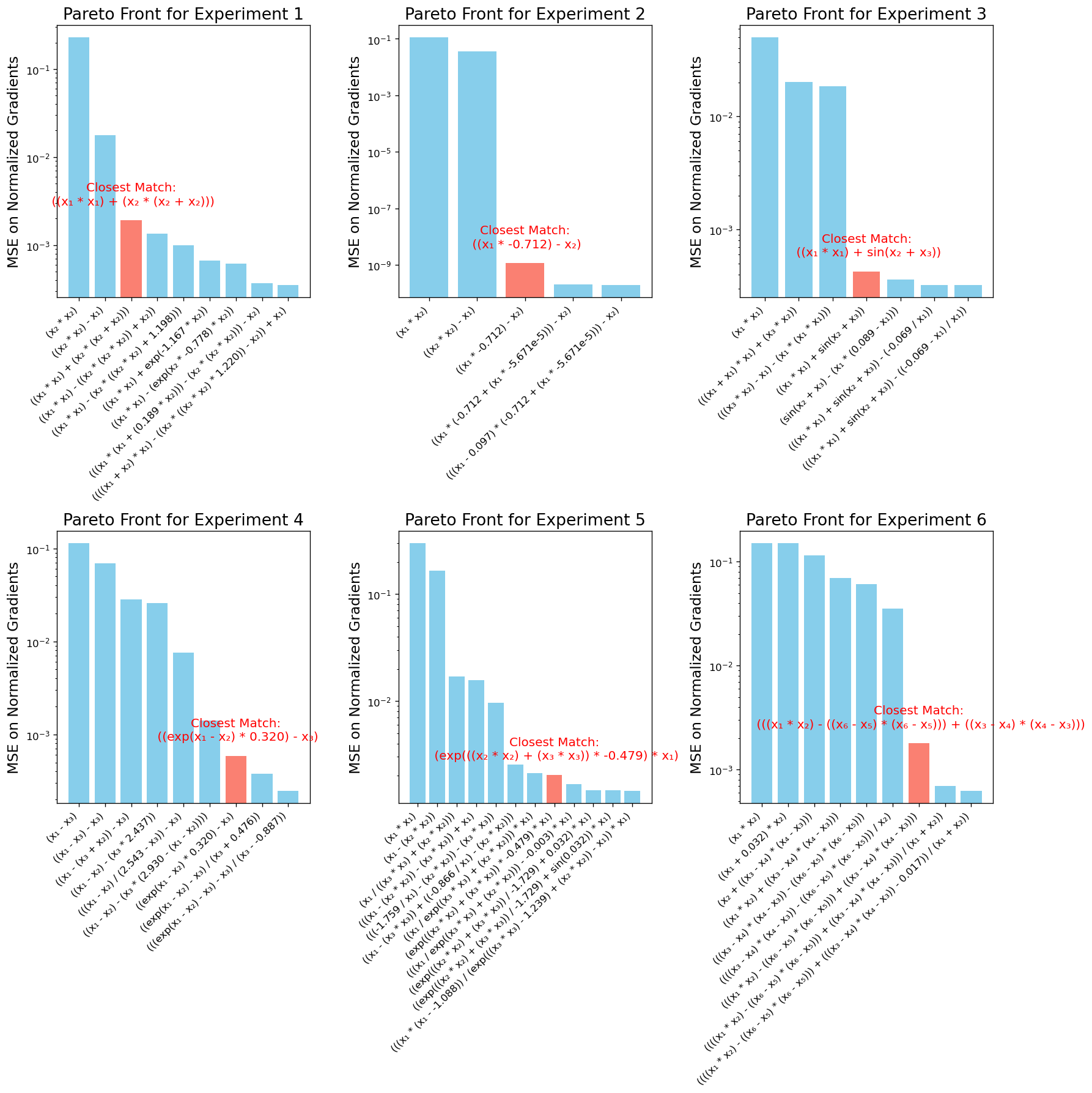}
    \caption{The results of fitting a symbolic model $T$ to the normalized gradients of the neural network are presented along the Pareto front. The Pareto front collects several possible results with decreasing Mean Square Error (MSE) and increasing complexity. The closest match to the true underlying function is often found at the point of steepest change of the Pareto front. }
    \label{fig:results}
\end{figure*} 

\subsection{Experiments 1-6, Interpreting Decision Function}

Applying the proposed interpretation framework to the data sets yields sets of formulas along the Pareto front. All formulas are equally valid and trade complexity for accuracy. The Pareto fronts for the experiments are collected in fig.~\ref{fig:results}.

The interpretation in experiment 1 is able to correctly recover the decision function $x_1^2+2x_2^2$. In experiment 2, the closest match is $-0.712x_1+x_2$ which seems incorrect at first glance. However, if we promote the transformation $\phi\rightarrow(\phi/0.712)^2$, we can map the current solution to a very good approximation of the true function $ x_1^2+ 2.81 x_1x_2  +1.97 x_2^2 $. In experiment 3 the algorithm correctly recovers $x_1^2+\sin(x_2+x_3)$. Similarly, promoting the transformation $\phi\rightarrow(\phi/0.32)^2$ in experiment 4 yields a very good approximation to the true function $\exp(x_1-x_2)-3.125 x_3$. In experiment 5, I recover the equation $x_1\exp(-0.479(x_2^2+x_3^2))$. However, in this experiment, the correct function is in a very flat region of the Pareto front. Lastly, experiment 6 fails to learn the division and instead replaces it with a subtraction $x_1x_2-((x_3-x_4)^2+(x_5-x_6)^2)$. Most likely the algorithm learns a Taylor approximation $1/(1+x)\approx1-x$ of the true function and removes any remaining constants since they do not contribute when calculating gradients. It is of note that in most experiments, the correct function can be found when the Pareto front exhibits the steepest change.

These experiments can be compared to direct symbolic classification in cases where there is no ambiguity in what high-level feature is responsible for the class boundary as in experiments 1-6. Symbolic classification is based on symbolic regression together with a loss function that enables a categorical output. The application of symbolic classification and the corresponding results can be found in appendix \ref{chapter:SymClass}. It recovers three out of the six decision formulas together with almost exact values for the biases/thresholds.

\subsection{Experiment 7, Symbolic Classification is not Neural Network Interpretation}
\label{chapter:Experiments2}

\begin{figure*}[h!]
    \centering
    \includegraphics[width=1\textwidth]{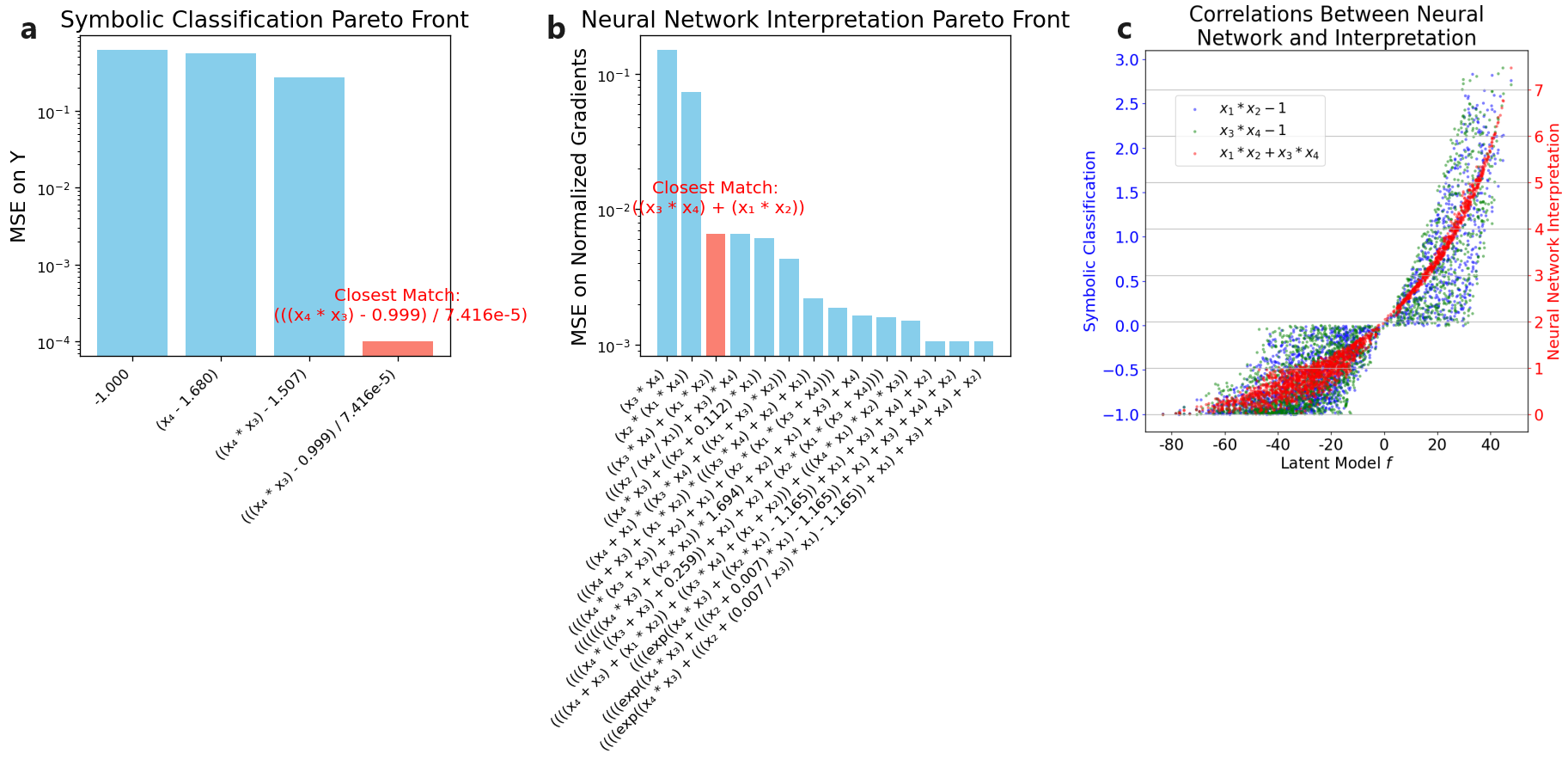}
    \caption{\textbf{a:} Results of fitting a symbolic classification model to the data of experiment 7. \textbf{b:} Interpretation of a neural network classifying the same data set. \textbf{c:} Empirical correlation between symbolic classification, the proposed interpretation method, and latent model $f$. Symbolic classification learns a different high-level feature than the neural network. The interpretation framework presented in this paper correctly interprets the neural network. }
    \label{fig:experiment7}
\end{figure*} 

Experiment 7 is different in the sense that the features are correlated and there are several possible concepts a model could learn to successfully classify the data. These correlations are common in machine learning data sets, eg when distinguishing between humans and dogs, one possible concept might be the presence of tails while another would be the color of the fur/skin. I train a neural network and interpret it similarly to the previous experiments. Further, I train a symbolic classification model to classify the data. In fig.~\ref{fig:experiment7}\textbf{a} one can see that the symbolic classification model selects one of the high-level features $x_1x_2$ to make its decision. The interpretation procedure reveals that the neural network bases its prediction on a combination of the two high-level features $x_1x_2+x_3x_4$, see fig.~\ref{fig:experiment7}\textbf{b}. In fig.~\ref{fig:experiment7}\textbf{c} it is shown that the neural network has a very low correlation with either of the single features and confirms a very strong correlation between the concept learned by the neural network and the result of the interpretation framework. In this section, one can see that symbolic classification does not necessarily provide an interpretation of the neural network, whereas the framework presented in this paper does.

\section{Discussion}
\label{chapter:Conclusion}
In this paper, I introduce a comprehensive framework for interpreting neural network classifiers that is equally capable of interpreting single neurons in hidden layers or latent spaces. The solution presents the learned concept as a closed-form function. The interpretation method is based on the idea of finding an intersection between the space of human-readable functions and all functions that contain the same information as a specific neuron in an artificial neural network. Through the interpretation procedure, it is possible to get rid of uninterpretable transformations that conceal the true function and focus on the simplest form of the underlying concept.

I demonstrate the power of the interpretation method in section \ref{chapter:Experiments} at the example of different data sets with two to six input variables each. In 4 cases the interpretation procedure recovers the exact functional form of the true decision function, while in 2 experiments the procedure finds a very good symbolic approximation. In one experiment the neural network has the freedom of choosing between two different high-level features to make its prediction. The interpretation method reveals that the neural network learns a combination of both features. 

Comparing these results with other works is impossible since most related works only replicate neural network regressors. As demonstrated in experiment 7, if there are multiple high-level concepts that can be employed to solve a specific task, replication is not interpretation. The most similar research articles manage to explain neural network regressors involving one variable \cite{alaa2019demystifying} or two variables \cite{Abroshan2023}. It is possible to compare the neural network interpretation technique to symbolic classification, see appendix \ref{chapter:SymClass}. Direct symbolic classification only finds three out of six closed-form expressions of the decision boundaries whereas gradient-based interpretation manages to find suitable solutions for all cases. There are multiple potential reasons for this performance difference. Neural networks are superior learners than genetic search algorithms. Symbolic search in the proposed interpretation framework is tasked with finding a closed-form expression extending deep into the full training manifold and hence experiences training signals from all data points. Direct symbolic classification only accumulates error signals from violations of the decision boundary. Neural networks are well suited to deal with noisy data such that the artificially created data of normalized gradients for the symbolic search steps could be essentially noise-free. Further, unlimited additional artificial data generation in algorithm \ref{algcollection}.2 would further enable better symbolic performance. 

The interpretation method can also be used to simplify the interpretation of neural network regressors. Consider for example 
\begin{align}
G(x_1,x_2)=&\frac{a}{\exp(b\ x_1+ c\sin (d x_2))+1}=\frac{a}{\exp(bz)+1}=g(z) \ ,\\& \text{with  } z= x_1+ c/b \sin (d x_2)
\end{align}
To find a symbolic expression of a neural network learning the function $G(x_1,x_2)$ in the context of regression, use the framework in this paper to find $z$, then prepare the symbolic regression problem for $g(z)$ and use your favorite symbolic regression algorithm to solve a simple one-dimensional symbolic regression problem.

There is of course the limitation of not being able to learn functions of a single variable since the interpretation method's goal is to reduce the concept learned by a neural network to a one-dimensional function. This also means one loses the numerical value of the threshold/bias of the output neuron. This is however not a problem, since solving a one-dimensional symbolic regression problem or fitting a threshold/bias is a minor challenge compared to the complexity of problems solved by the presented interpretation framework.

% A central open question remains: why does a neuron contain the concept when it evaluates the function far away from the decision boundary? In my intuition, there is only a very weak training signal induced by inputs that are already classified correctly.

When applying the interpretation algorithm to practical problems it might be useful to sort out equations from the Pareto Front that violate dimensional constraints. Similarly, dimensional analysis could also be used inside the symbolic search algorithm itself. Further, one might have some inductive bias in mind that is better represented through other function sets. There is no need to use symbolic trees to represent the solution, any suitable differentiable function space can do. Lastly, this interpretation method is more useful in bottleneck layers and when neurons are disentangled, since the framework cannot capture information distributed among multiple neurons.

This article is accompanied by the article "Closed-Form Interpretation of Neural Network Latent Spaces with Symbolic Gradients" \cite{https://doi.org/10.48550/arxiv.2409.05305}, where we show how to interpret latent spaces of neural networks in closed form.

The code used for this project can be found at \url{https://github.com/sjwetzel/PublicSymbolicNNInterpretation}

\section{Acknowledgements}
I thank the National Research Council of Canada for their partnership with Perimeter on the PIQuIL. Research at Perimeter Institute is supported in part by the Government of Canada through the Department of Innovation, Science and Economic Development Canada and by the Province of Ontario through the Ministry of Economic Development, Job Creation and Trade. This work was supported by Mitacs through the Mitacs Accelerate program. I also thank the PySR team, especially Miles Cranmer, for their helpful comments on github.
\\

\newpage
\appendix

\section{Symbolic Classification}
\label{chapter:SymClass}

\begin{figure*}[h!]
    \centering
    \includegraphics[width=1\textwidth]{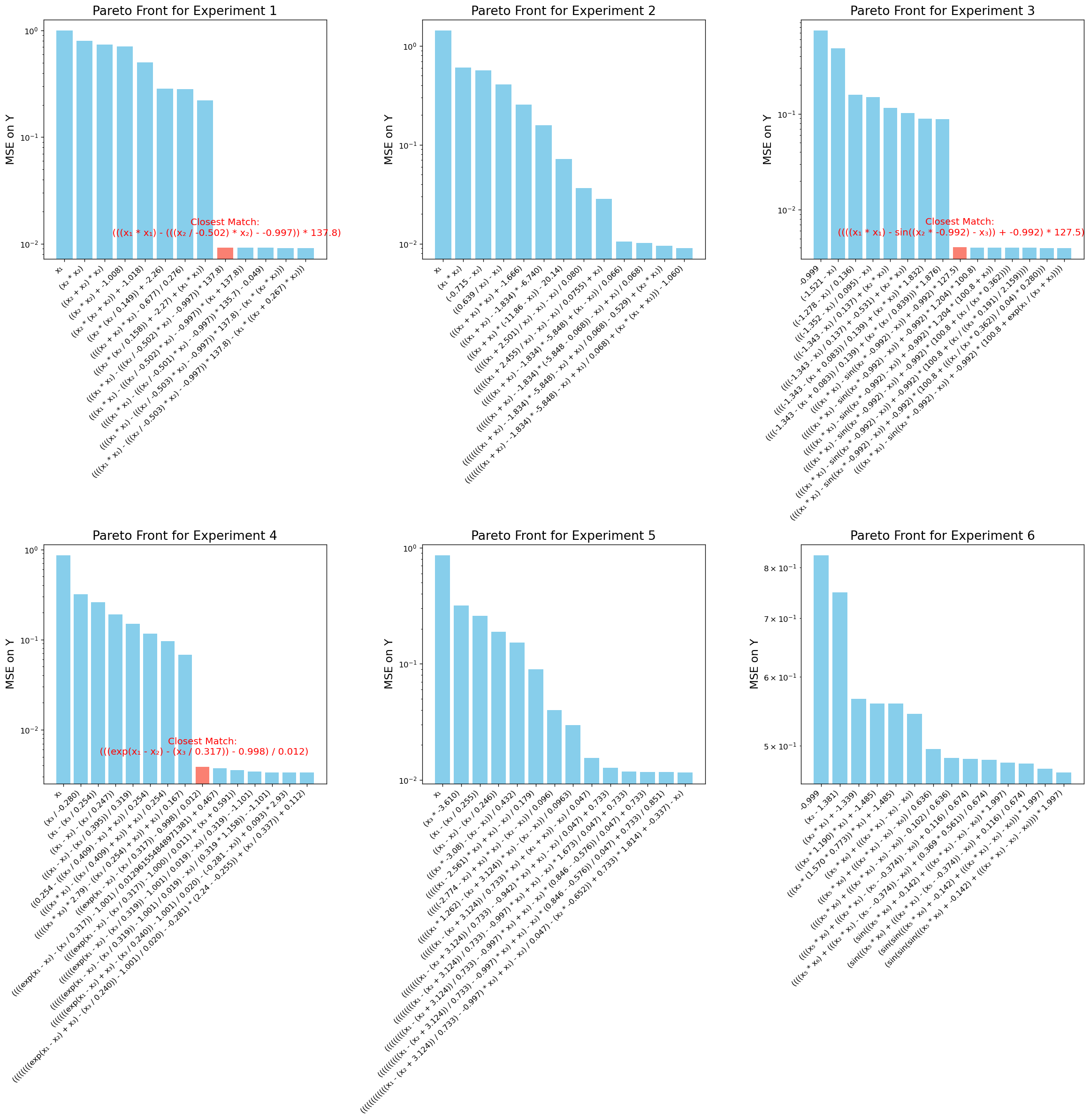}
    \caption{The results of fitting a symbolic classification model $T$ to six experiments. The Pareto front collects several possible results with decreasing Mean Square Error (MSE) and increasing complexity. The closest match to the true underlying function is often found at the point of steepest change of the Pareto front.}
    \label{fig:symclassresults}
\end{figure*} 

The most closely related technique to learn what neural network classifiers learn is symbolic classification. However, it is important to note that symbolic classification is not an interpretation technique, it can only provide equivalent results if there is only one possible high-level concept that a neural network could learn in order to make its predictions. Symbolic classification is based on symbolic regression with hinge/margin loss function, more precisely, the loss function I use is 
\begin{align}
\mathcal{L}(F(X),Y)=\frac{1}{n}\sum_{i=1}^n \max (0,1-(y_iF(\mathbf{x}_i)))
\end{align}
I perform symbolic classification on experiments 1-6 with the same symbolic regression algorithm and hyperparameters as the experiments in the main body of this paper. The model is trained using the following building blocks: input variables $x_1,...,x_n$, floating-point constants, binary operators: +, -, *, /, unary operators: sin, exp. The hyperparameters are a batch size of 25, a maximum number of epochs of 200, an early stop callback, and a maximum tree size of 30 nodes.

The Pareto fronts describing the results can be found in fig.~\ref{fig:symclassresults}. In three experiments 1, 3, 4 the symbolic classification algorithm finds the exact decision boundary, while in experiments 2, 5, 6 the algorithm fails to discover anything meaningful. Symbolic classification has the advantage of also recovering the threshold/bias of the last layer in contrast to the interpretation framework presented in this manuscript. However, symbolic regression adds increased complexity to its algorithm because it spends nodes on the exact determination of the bias/threshold. Further, the gradient-based interpretation framework has the freedom to choose from a larger function set and can thus provide more simplifications in its solutions.
\\

% \section{Declarations}
% This work was supported by Mitacs and Homes Plus Magazine Inc. through the Mitacs Accelerate program. The code supporting this publication is available at \cite{githublink}. The data sets used in this work are described in \ref{sec:data}, they can either be found online at \cite{datasets} or generated from equations. I consent to publishing all code and data.

% \noindent *Conflicts of interest/Competing interests - not applicable\\
% *Ethics approval - not applicable\\
% *Authors' contributions - S.J.W did all the work\\
% *Funding - Mitacs\\
% *Consent to participate - I consent\\
% *Consent for publication - I consent\\
% *Availability of data and material - https://github.com/sjwetzel/PublicSymbolicNNInterpretation\\
% *Code availability - https://github.com/sjwetzel/PublicSymbolicNNInterpretation
\bibliography{library}
\bibliographystyle{unsrturl}

\newpage

\appendix
\end{document}